\documentclass[letterpaper, 10 pt, conference]{ieeeconf}  

\IEEEoverridecommandlockouts                              

\usepackage{graphics} 
\usepackage{epsfig} 
\usepackage{mathptmx} 
\usepackage{times} 
\usepackage{amsmath} 
\usepackage{amssymb}  
\usepackage{mathtools}
\usepackage{multirow}
\newcommand{\RNum}[1]{\uppercase\expandafter{\romannumeral #1\relax}}
\usepackage{hyperref}
\hypersetup{colorlinks, linkcolor=red}

\title{\LARGE \bf
Slip Detection with Combined Tactile and Visual Information
}

\author{Jianhua Li$^{1*}$, Siyuan Dong$^{2*}$ and Edward Adelson$^{3}$
\thanks{$^{1}$Jianhua Li is with Computer Science and Artificial Intelligence Laboratory (CSAIL), MIT, Cambridge, MA 02139, USA
        {\tt\small jianhuali@csail.mit.edu}}%
\thanks{$^{2}$Siyuan Dong is with the Department of Electrical Engineering \& Computer Science, and CSAIL, MIT, Cambridge, MA 02139, USA
        {\tt\small sydong@mit.edu}}%
\thanks{$^{3}$Edward Adelson is with faculty of Department of Brain and Cognitive Sciences, and CSAIL, MIT, Cambridge, MA 02139, USA
        {\tt\small adelson@csail.mit.edu}}%
\thanks{*Jianhua Li and Siyuan Dong contributed equally to this work.}}

\begin{document}

\maketitle
\thispagestyle{empty}
\pagestyle{empty}

\begin{abstract}

Slip detection plays a vital role in robotic manipulation and it has long been a challenging problem in the robotic community. In this paper, we propose a new method based on deep neural network (DNN) to detect slip. The training data is acquired by a GelSight tactile sensor and a camera mounted on a gripper when we use a robot arm to grasp and lift 94 daily objects with different grasping forces and grasping positions. The DNN is trained to classify whether a slip occurred or not. To evaluate the performance of the DNN, we test 10 unseen objects in 152 grasps. A detection accuracy as high as 88.03\% is achieved. It is anticipated that the accuracy can be further improved with a larger dataset. This method is beneficial for robots to make stable grasps, which can be widely applied to automatic force control, grasping strategy selection and fine manipulation.

\end{abstract}


\section{INTRODUCTION}
A contact state feedback to the control system is important for many robotic manipulation tasks. Slip, a common losing-contact state, occurs when the grasp is executed with insufficient force or improper grasping strategy. Detecting slip and incipient slip can assist robots to automatically adjust the grasping force and choose an appropriate plan of action. Slip detection has a lot of applications in both industrial and service robots. It is unsurprising that people have developed many tactile sensors and methods to do slip detection in the past decades. A good review paper can be found here~\cite{francomano2013artificial}. However, to the best of our knowledge, there is no good commercialized tactile sensor that can detect slip with high accuracy. 

\begin{figure}[thpb]
	\centering
    \includegraphics[width=0.40\textwidth]{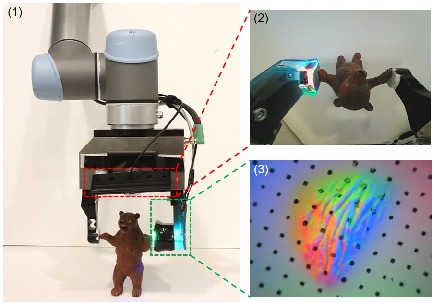}
    \vspace{-5pt}
	\caption{(1) The experiment setup: UR5 robot arm and WSG-50 parallel gripper. One finger of the gripper is replaced with GelSight sensor. An external camera (normal webcam) is mounted on the top of the gripper. The gripper is holding a toy bear. (2) The image captured by the  external camera. (3) Image from GelSight sensor.}
	\label{fig:object_in_gripper}
\end{figure}

The GelSight sensor is an optical tactile sensor developed in 2009~\cite{GelSight2009}. Among different kinds of tactile sensors, the GelSight sensor has the advantage of high spatial resolution ($640 \times 480$ pixels, 0.024 mm/pixel) since it uses a camera to image the contact surface. The movement of an object against the contact surface can be captured with a spatial resolution higher than that of human fingers. In addition, the displacement of the black markers on the sensor surface (Figure~\ref{fig:object_in_gripper}(3)) can be used to infer the normal, tangential and torsional forces. Slip happens when the texture of the object moves against the markers on the sensor surface. For objects without texture, slip can be detected by analyzing the motion of markers in the contact area. Therefore the GelSight sensor is capable of detecting slip. In our previous work~\cite{GelSight_greensensor}, we measured the relative motion between the texture of the object and the markers and set a slip threshold. When the relative displacement is above the threshold, we concluded that slip happened. The method is effective in most conditions and we achieved a detecting accuracy as high as 71\%. However, the challenging problem of this method is to find the threshold that performs well for objects with various shapes and weights. Furthermore, when the object barely touches the surface, the signal from the Gelsight is too weak to be used for slip detection. 

Humans estimate the grasping force before lifting an unknown object and adjust the force accordingly. When slip happens, they naturally increase the force until the grasp is stable. The process is very short and the object is only slightly lifted. The lifting process is necessary since both the weight and the relative friction coefficient between the object and human fingers are important for choosing a proper grasping force. In this paper we proposed a new method to detect slip, mimicking the strategy adopted by humans. We employs a DNN with images captured by a GelSight sensor and an external camera mounted on the gripper. The  external camera and the GelSight sensor act as human fingers, and the DNN is like the human brain. We perform grasping and lifting experiments with different forces and grasping positions on 84 daily objects for 1102 times to train the neural network. We use 10 new objects to test the DNN and achieve 88.03\% accuracy. Our dataset is available at \url{https://www.dropbox.com/sh/i39pjxfqiwhbu1n/AAD5FNjk-Nt28UZ8lsVuVd4ja?dl=0}

This paper is organized as follows: in Section II, related work of slip detection and GelSight sensor is explained. In Section III, we describe the architecture of the DNN and training specifications. In Section IV, we show the experimental results and discuss what the neural network learns from the data to detect slip. Finally, we summarize the contributions of this paper and discuss potential applications of this method.

\section{RELATED WORK}

\subsection{Slip detection}
Slip detection is very important for robotic manipulation. In the past decades, people developed various tactile sensors to detect slip by sensing physical signals, like vibration, thermal change, normal and tangential force, acceleration and relative motion between fingertips and the object. Cotton \textit{et al}.~\cite{cotton2007novel} proposed a thick-film piezoelectric tactile sensor to detect slip by sensing the vibration, which could be used as the fingertip of a dexterous hand. In 2000, Melchiorri~\cite{melchiorri2000slip} proposed a method, which measured the ratio of normal and shear components of the contact force and compared it to the frictional coefficient of the surface, to detect both translational and rotational slip by use of a force/torque and a tactile matrix sensor. However, the frictional coefficient of the contact surface needs to be measured  in advance. Accoto \textit{et al}.~\cite{accoto2012slip} developed a thermal slip microsensor in 2012, which is a planar gold microheater that measures temperature change in the contact surface. When slip happens, there is heat flow from the sensor to the object that will be the trigger signal. Unfortunately, it is difficult for thermal sensors to distinguish slip signal and contact signal. 

The method of measuring relative motion between finger tips and the object is mainly utilized in optical based tactile sensors. In 2002, Hosoda~\textit{et al}.~\cite{hosoda2002internal} built a robot hand with anthropomorphic fingertips, which was equipped with vision and tactile sensors. By training a neural network with vision and tactile data, a primitive representation of slip phenomenon was produced. The idea of combining vision and tactile information is quite similar to our work. However, their method could not distinguish slip and pressure change and a real grasping experiment was not implemented. Yuan \textit{et al}.~\cite{GelSightShear} proposed that by analyzing the contact condition the GelSight sensor can be used to detect slip. It is demonstrated that slip started happening from the peripheral area of the contact surface in their experiment, where the motion of the sensor surface can be visualized by the markers on it. So a larger displacement of the central marker compared to that in the peripheral area indicated slip. However, they didn't implement any robot experiments either and the objects chosen to test had very little texture on the surface. 

Calandra \textit{et al}. ~\cite{slip_UCB} used two GelSight sensors along with an RGB camera to predict slip when picking up daily objects with a robot arm. (Note that slip prediction, which occurs at the initial grasp, is different from slip detection, which occurs as the object is being lifted). They used end-to-end deep learning and achieved prediction accuracy of 94\%.
 
Dong \textit{et al}.~\cite{GelSight_greensensor} detected slip using a GelSight sensor by combining the strategy of ~\cite{GelSightShear} with the method that directly measured the relative displacement between the marker and texture for textured objects. They tested the new method by implementing a robotic grasping and lifting experiment with 37 daily objects and 315 grasping times and got 71\% accuracy. The principle of the new method was very intuitive, but involves lots of threshold tuning problems for different objects. In addition, for objects with smooth surface and small contact area with the sensor, it is challenging  to detect slip of the object by only seeing the GelSight image sequences. Therefore, in this paper, we add an external camera mounted on the side of the gripper to give another cue for slip detection. 

\subsection{GelSight sensor}
A GelSight sensor is a camera-based tactile sensor, which was designed to measure the fine geometry profile of the contact surface~\cite{GelSight2009,GelSight2011}. GelSight sensor is mainly made up of three components: soft silicone gel with a reflective membrane, three color LEDs and a regular webcam. The silicone gel is illuminated by red, green and blue (R,G,B) light with three different directions. The deformation of the gel when contact happens is captured by the camera on top of the gel. The R, G, B values in the captured image can be used to infer the depth map of the contact surface. Li \textit{et al}.~\cite{GelSightUSB} designed a fingertip GelSight sensor with much smaller volume, which can be used as the finger of a robot gripper. The sensor was attached to a Baxter robot hand and completed a USB insertion task. Yuan \textit{et al}.~\cite{GelSightShear} further improved the sensor by adding markers on the gel surface. The marker motion indicates the normal, shear, torsional force on the contact surface and can even detect incipient slip. Dong \textit{et al}.~\cite{GelSight_greensensor} recently developed a new version of GelSight fingertip sensor that measures the geometry of contact surface more accurately. The new sensor was used to detect slip by analyzing the relative motion between object and gel surface and the marker motion around the contacted area. Compared to other tactile sensors, GelSight has much higher spatial resolution ($640\times 480$ pixels) so that any tiny change in the contact surface can be observed. A large amount of 2D data can be easily taken from a GelSight sensor.
\begin{figure*}[h]
	\centering
       \includegraphics[width=0.88\textwidth]{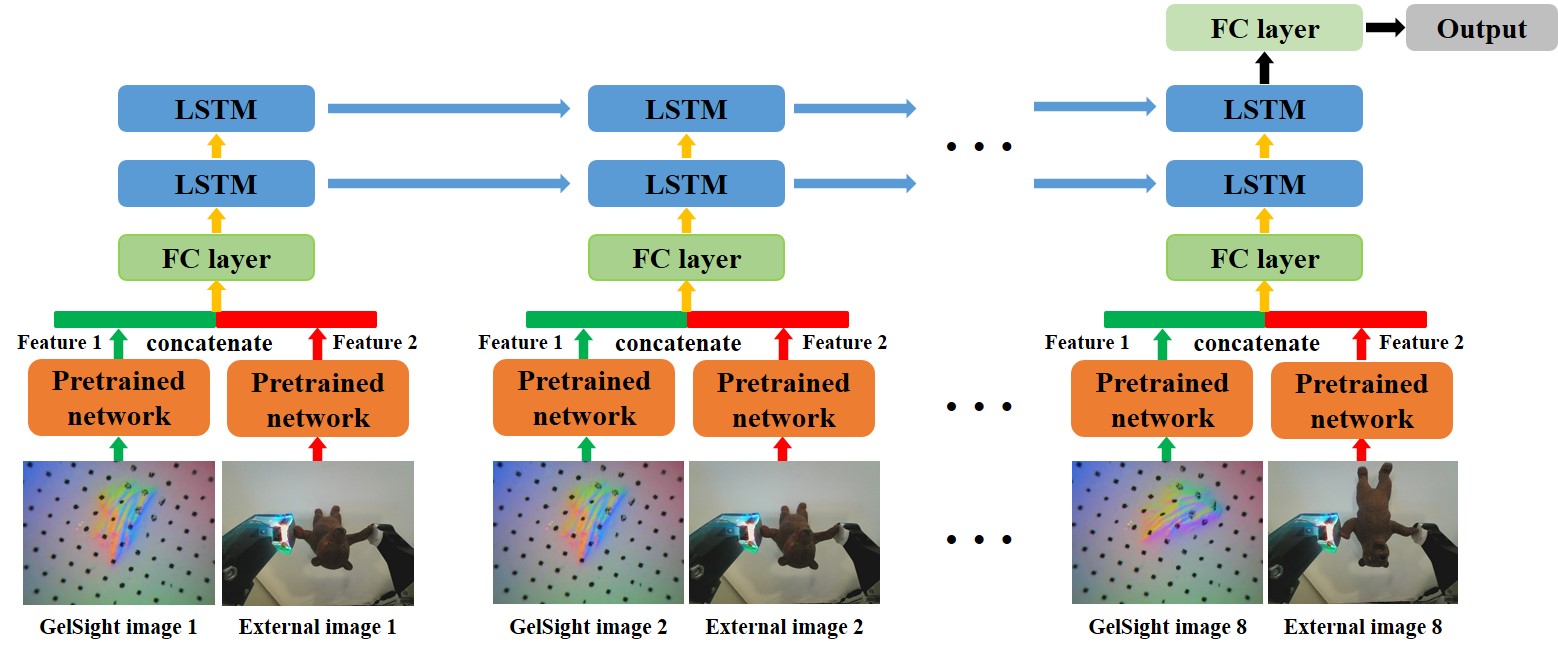}
    \vspace{-5pt}
    \caption{The diagram of the DNN architecture.}
    \label{fig:DNN}
\end{figure*}

\begin{figure}[t]
	\centering
       \includegraphics[width=0.49\textwidth]{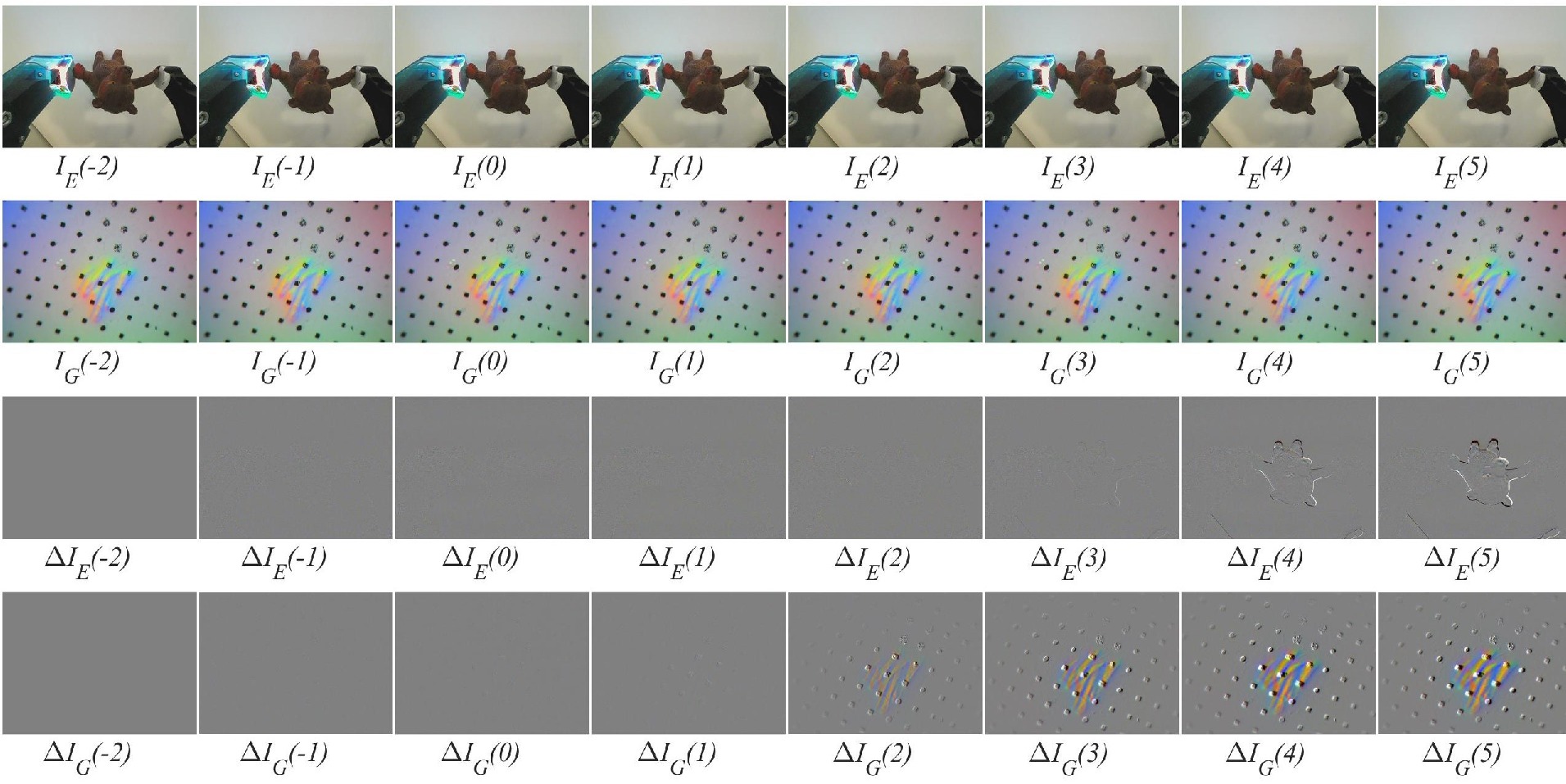}
    \vspace{-15pt}
    \caption{Image sequence captured when the robot arm lifts a toy bear. At time $T(0)$, the robot arm starts to lift the toy bear. The images $I_E(t)$ ($t=-2,-1,0,\hdots$) in the upper row are taken by the external camera and the ones $I_G(t)$ in the second row are from a GelSight sensor. The images $\Delta I_E(t)$ and $\Delta I_G(t)$ in the third and forth row are the image difference between the images captured at time $T(t)$ and $T(-2)$, and computed as $\Delta I_E(t) = 128 + I_E(t)-I_E(-2)$ and $\Delta I_G(t) = 128 + I_G(t)-I_G(-2)$ respectively. An offset (128) is added to the image difference to ensure a positive value in each pixel.}
    \label{fig:data}
\end{figure}

\section{MODEL DESCRIPTION}

In this section, we introduce the architecture we used for the DNN, which is shown in Figure~\ref{fig:DNN}. For the input, the length of the image sequence we choose is eight frames. The output of the DNN is the probability of slip happening. The DNN mainly contains two parts: 1) Convolutional Neural Network (CNN) to extract features of each image and 2) Recurrent Neural Network (RNN) to compare the feature sequences and make a decision. For the CNN part, a pretrained neural network, such as VGG-16 network~\cite{VGG} which was trained on ImageNet~\cite{deng2009imagenet}, is used. We pair one GelSight image with one external-camera image, which are almost captured at the same time. Since we are using eight images as a sequence, eight pairs of images from two cameras are grouped as input. The 16 images are fed into the pretrained network separately, yielding 16 rows of features. Then for each pair, the two output features are concatenated into one new row. To select useful features, the pretrained network is followed by a fully-connected (FC) layer and the features are shrunk into 64 features. For the RNN part, we choose to use Long Short Term Memory networks (LSTM)~\cite{LSTM}, which is the most popular RNN architecture that can connect long term information to the present task. LSTM has been widely used to process video data~\cite{srivastava2015unsupervised,yue2015beyond}, which is also similar to our task. The 8 rows of features, which contain information of the relative motion of the object with GelSight sensor, are filled into two layers of LSTM with 64 memory units. Then a final FC layer with two outputs is added to classify a stable grasp or slip. In order to avoid overfitting, a dropout layer with the keep-probability of 50\% is added after the FC layer and a dropout layer with the keep-probability of 80\% is added after each LSTM layer. 

It is very difficult to train such a deep neural network, especially with a relatively small dataset. So we fix the parameters in pretrained network and only train the rest of the DNN. For the trainable layers, we use random initialization and cross-entropy as the loss function. Adam optimizer~\cite{kingma2014adam} with $5\times10^{-4}$ learning rate is used in the training process. We build the neural network by using TensorFlow package, and train it on a Nvidia GeForce Titan X GPU with 12 GB memory. The batch size we choose is 160.

\section{EXPERIMENTAL RESULT}
\subsection{Experimental setup and data collection}

\begin{figure*}[tpb]
	\centering
	\includegraphics[width=\linewidth]{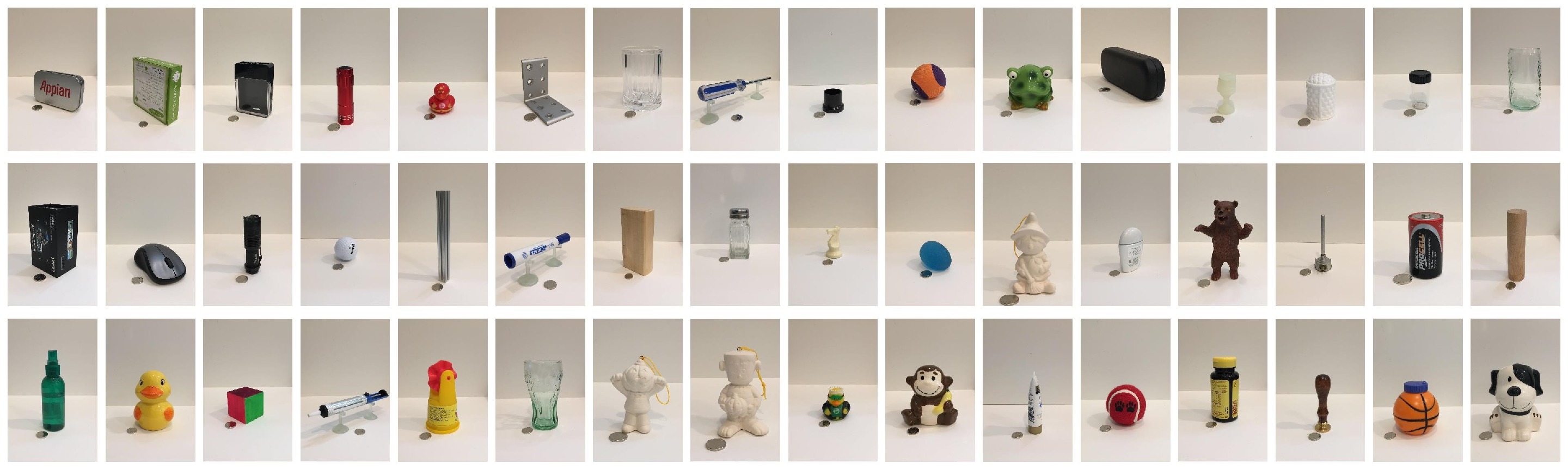}
    \vspace{-15pt}
	\caption{Some selected objects that are used to train the model.}
	\label{fig:object_figure}
\end{figure*}

We conduct the grasping experiment with a 6-DOFs UR5 robot arm equipped with a WSG 50 parallel gripper. One finger of the gripper is replaced by a GelSight sensor and a normal webcam is mounted on the side of the gripper. The whole setup is shown in Figure~\ref{fig:object_in_gripper}(1). The maximum opening distance of the gripper is 110.0 mm, which is, however, reduced to 80.0 mm because of the width of the GelSight sensor. The lifting speed of the robot arm is set to 40.0 mm/s and the gripping speed of the gripper is 10.0 mm/s. 


We perform the grasping and lifting experiments on 84 daily objects. Limited by the space, only some selected objects are shown in Figure~\ref{fig:object_figure}. Generally, the objects have different sizes, shapes, surface textures, material and weights. The widths of the objects are chosen to be smaller than the maximum opening distance of the gripper. In the experiment, firstly, a barely touching distance of the gripper for each object is measured automatically with the feedback of the GelSight sensor. Around this distance, different gripping distances are chosen to balance the number of stable grasping cases and slip cases. Each object is grasped and then lifted up slowly for 20.0 mm, which is empirically enough to detect the stability of the grasp. During the lifting process, the data are collected by the GelSight sensor and external camera with 20 Hz and 640$\times$480 resolution. 

As mentioned in Section \RNum{3}, a sequence of 8 successive images are used as one set of input for the DNN. To select the image sequence we are interested in from the raw data, we set the time when the arm starts lifting as a reference time $T(0)$. The first image is chosen to be the image captured at time $T(-2)$, which is two frames before the time $T(0)$, to ensure that an image when the object is statically grasped gets included in the input sequence. Then the rest of the 7 images are successive images starting from time $T(0)$. 

We conduct 1102 grasping and lifting experiments in total to train our model, which is relatively a small dataset for a neural network method. Figure~\ref{fig:data} shows an example of the image sequence of a toy bear being lifted. The first row is image sequence ($I_E(t)$, $t=-2,-1,0,\hdots$) from the external camera, and the second row is the sequence ($I_G(t)$) captured by the GelSight sensor. To increase the size of our dataset, we use a sliding window to select the 7 successive frames in different places of the whole image sequence with the stride of 1 frame, and the frames at $T(-1)$, $T(1)$, $T(2)$ and $T(3)$ are used as the beginning of the 4 newly generated image sequences. By using this method, we generate 5 pieces of data from one image sequence. Overall, we generate 5510 pieces of data,  85\% of which were used for training and the other 15\% were used for validation. In the dataset, both translational and rotational slip are included and the border cases (incipient slip) are categorized as slip, since it also indicates an unstable grasp.

\begin{figure*}[tpb]
	\centering
	\includegraphics[width=0.95\linewidth]{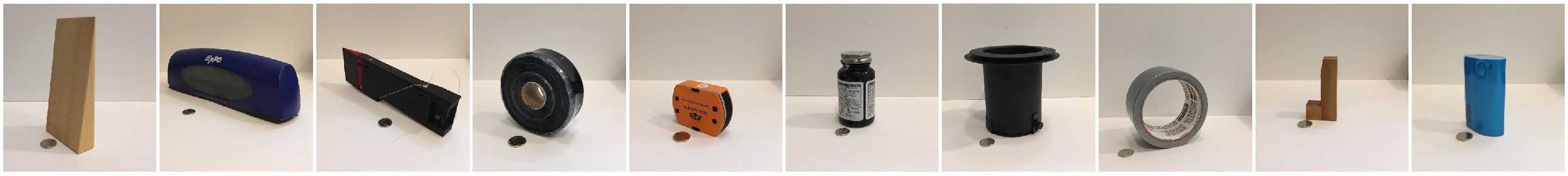}
    \vspace{-5pt}
	\caption{Ten objects are used to evaluate the the performance of our model. These objects were not seen during training, and differ significantly in weight, color, shape, dimension, and material from the training objects.}
	\label{fig:test_objects_figure}
\end{figure*}
\vspace*{-5pt}

\subsection{Slip detection results} 

To evaluate the performance of the models, we use 10 objects that are not shown in the training data as test objects. As shown in Figure~\ref{fig:test_objects_figure}, the ten objects also have different shapes, materials and weights. Each object is grasped and lifted around 15 times, and the number of successful cases and failures cases are balanced. In total, the test objects are grasped with 152 grasps and we acquire 760 image sequences as the test data. In this section, the slip detection results of our model are presented and also some comparisons of different parameters are discussed. 

We study and optimize the parameters of the DNN model in three different aspects: data formats of the input, types of pretrained CNN and combination of data sources. 


\textbf{Data format} Because of the limited size of our dataset, the parameters of the CNN are fixed to be the pretrained weights on ImageNet to make the neural network trainable. The pretrained weights with ImageNet is to classify different objects, which implies that the position information of the target object in the image needs be ignored by the network. However, the relative position of the object in the image is the most useful information in the slip detection task. To overcome the problem, we create another image sequence by computing image difference between the images captured at time $T(t)$ and $T(-2)$, and define them as $\Delta I_E(t) = 128 + I_E(t)-I_E(-2)$ and $\Delta I_G(t) = 128 + I_G(t)-I_G(-2)$ respectively. The added offset (128)  to the image difference is to ensure a positive value in each pixel. As shown in Figure~\ref{fig:data} third and fourth row, the motion of the toy bear captured by the external camera and GelSight sensor is highlighted in the new sequences. The two different inputs are used to train multiple models. 

\textbf{Pretrained CNN} Another parameter we tweak is the pretrained CNN model. Since the CNN is supposed to extract useful features that are helpful to the slip detection task, choosing an appropriate CNN model is important. Three popular CNN models have been tried here: VGG-16, VGG-19 and Inception-V3~\cite{szegedy2016inception}. All of them are used with pretrained weights on ImageNet. For the VGG-16 and VGG-19, we extract features from their layers \textquotedblleft$fc7$" which have 4096 descriptions of an image. And for the Inception-V3, we get features from its layer \textquotedblleft$pool\_3:0$", which is a tensor containing the next-to-last layer containing 2048 descriptions of an image.

\textbf{Data source} There are three different combinations of the data sources: tactile-only, vision-only and tactile-vision. We train three models by using each of them as input to find out the best data source. In addition, in our previous work~\cite{GelSight_greensensor}, only GelSight data was used to detect slip. We also do a comparison between the DNN training on GelSight data only and the method in~\cite{GelSight_greensensor}.

The test results of the models are summarized in Table~\ref{tab:result_tab}. It is obvious that the union of tactile and vision data gives the best test accuracy regardless of the CNN type and input data format, which demonstrates that vision and tactile sensors are mutually promoted in the slip detection task. With raw image sequence as the input, the Inception-V3 model gives 88.03\% test accuracy by using vision and tactile data, which is significantly higher that those of VGG-16 (82.11\%) and VGG-19 (78.55\%). Considering the computational speed, a smaller DNN model is preferred in the real robotic experiments. So Inception-V3, which is a much smaller model than the other two models, is the best fit in this circumstance. Interestingly, the test accuracies of the models, which only use the vision raw image sequence as the input, are as low as around 50\%, which implies the neural network doesn't learn any useful information. As analyzed above, one possible reason is the object position information in the image sequence gets reduced or is lost by the CNN network pretrained on ImageNet. In contrast, by using the image difference input to train the model, three much higher accuracies are achieved with all of the three DNN vision-only models. Since GelSight images look very different from natural images, the features extracted by the pretrained CNN can also be quite different. That probably explains why the performance of the models that only use the GelSight images as input are not much affected by the different input formats. Among all of the models, the best test accuracy (88.03\%) is achieved by using raw image input, Inception-V3 CNN and tactile-vision data sources. 



\begin{table}[tpb]
\caption{Experimental results. A sequence of 8 continuous frames are used as the input.}
\vspace{-10pt}
\label{tab:result_tab}
\begin{center}
\begin{tabular}{|c|c|c|c|c|} 
\hline
input& feature & Tactile-vision & Tactile & vision \\
\hline
\multirow{3}{*}{\parbox{1.2cm}{raw image sequence}}  & VGG16-fc7 & 82.11\% & 81.84\%& 55.13\% \\ \cline{2-5}
& VGG19-fc7 & 78.55\% & 75.39\%& 55.39\% \\ \cline{2-5}
& Inception-V3 & \textbf{88.03\%} & 82.24\% & 53.68\% \\ \cline{2-5}
\hline
\multirow{3}{*}{\parbox{1.2cm}{image difference sequence} }  & VGG16-fc7 & \textbf{87.76\%} & 74.87\% & 79.74\% \\ \cline{2-5}
& VGG19-fc7 & 85.53\% & 76.18\% & 77.37\% \\ \cline{2-5}
& Inception-V3 & 83.68\% & 78.82\% & 80.92\% \\
\hline
\end{tabular}
\end{center}
\end{table}

\begin{table}[tpb]
\centering
\caption{Test results of DNN with different lengths of input sequence using tactile and vision images}
\vspace{-15pt}
\label{tab:input_length}
\begin{center}
\begin{tabular}{|c|c|c|c|c|} 
\hline
\multirow{2}{*}{Model parameter} & \multicolumn{4}{|c|}{input length}\\ \cline{2-5}

& 6 & 7 & 8 & 9  \\ \cline{2-5}
\hline
{raw image, Inception-V3}  & 86.71\% & 83.95\% & \textbf{88.03\%} & 86.45\% 
\\ \hline 
{image difference, VGG-16}  & 84.08\% & 85.79\% & \textbf{87.76\%}  & 86.58\% \\ \hline  
{slip detection in ~\cite{GelSight_greensensor}}  & 53.28\% & 63.81\% & 78.28\% & \textbf{82.24\%}  \\ \hline  
\end{tabular}
\end{center}
\end{table}
After identifying the best parameters of CNN type and data source, the most appropriate input length is inspected. A short input length will dramatically increase the processing speed, but may miss useful information. We use different lengths of input (6, 7, 8, 9 frames) to train the model and the results are summarized in Table~\ref{tab:input_length}. With both model parameters, the length of 8 frames gives the highest test accuracy, but the results of 6 frames are also comparable. In addition, we process the same test data (only GelSight images) by using the slip detection method in~\cite{GelSight_greensensor}. The results with different input lengths, summarized in Table~\ref{tab:input_length}, show that our models give much better results in all of the cases. Especially, the dramatical difference between the two 6-frames-input results demonstrates that our model can detect slip signals earlier than that in~\cite{GelSight_greensensor}. 

\subsection{Experimental analysis} 

\begin{figure}[h]
	\centering
	\includegraphics[width=0.9\linewidth]{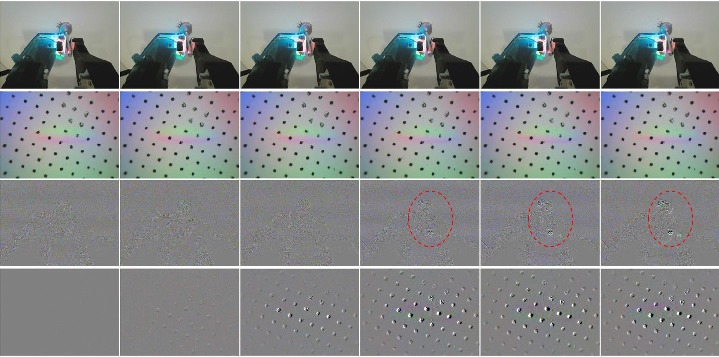}
    \vspace{-5pt}
	\caption{The first and second rows are image sequences captured by external camera and GelSight sensor respectively when a rheostat gets lifted; the third and fourth rows are image differences between the current image and the first image of the raw image sequences.}
	\label{fig:slipper_object}
\end{figure}

According to the experimental results, the model combining the GelSight sensor and external camera gives much better accuracy than that of any single source model, including GelSight only model. Theoretically, the GelSight sensor is able to provide the geometrical information of the contact surface, and the marker array on its surface acts as a reference when measuring the relative displacement between the object and sensor surface. All of the characteristics of the GelSight sensor make it an excellent tool to detect slip, however, in certain circumstances, it could be very difficult to judge whether slip happens or not by only observing the GelSight images. One example selected from the training data is shown in Figure~\ref{fig:slipper_object} and this is a slip case when a cylinder-shape rheostat with very smooth surface is lifted. The GelSight image sequence in the second row keeps repeating the same shape, which gives the illusion that the object is stable in the hand. The image difference row of GelSight images (the fourth row) only changes a bit when the sensor surface gets stretched at the lifting moment, but stays the same afterwards. So when the objects with smooth surface and same shape and appearance slip, the relative displacement cannot be reflected  from the GelSight images and the stretch force is too subtle to change the marker motion in the sensor surface. This makes the slip almost impossible to detect by the GelSight sensor. However, since the rheostat is not moving with the robot hand, its positions in the external camera image sequence change, which can be clearly seen in the last three images of the third row image sequence (highlighted by a red circle). This example well explains why the tactile-vision model performs better than other single source models.

\begin{figure}[h]
	\centering
	\includegraphics[width=0.9\linewidth]{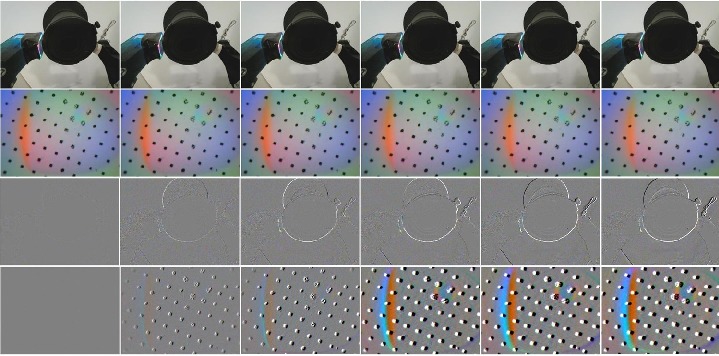}
    \vspace{-5pt}
	\caption{The first and second rows are image sequences captured by external camera and GelSight sensor respectively when a heavy metal object gets lifted; the third and fourth rows are image differences between the current image and the first image of the raw image sequences.}
	\label{fig:shear_example}
\end{figure}

One of the most difficult parts of the slip detection task is to distinguish whether there is relative motion between object and sensor surface. Especially for GelSight sensor, the soft gel surface will get stretched by the shear force when an object gets lifted. Figure~\ref{fig:shear_example} shows the image sequences when a heavy metal object gets lifted. From the two image difference rows (third and fourth) in the figure, it is obvious that the object is moving downwards in the lifting process. However, it is a stable grasp since there is no relative motion between the object texture and the black markers on the sensor surface shown in GelSight images. The motion of the object results from the stretch of the gel surface. Our neural network successfully classifies the grasp as stable, which demonstrates that the network learns some effective strategies to distinguish between slip and non-slip.

\section{CONCLUSIONS}
To summarize, we present a slip detection method by using GelSight tactile sensor and an external camera mounted on the side of the gripper without any pre-knowledge of the physical parameters of the objects. By using the image sequences captured by the two sensors when the objects begin to be lifted, we train a DNN to classify a grasp to be stable or not. We perform more than 1200 grasp experiments with 94 objects in total; 84 of them are used for training the network and we achieve 88.03\% detection accuracy when testing the rest of the 10 unseen objects. We also compare our model with other models, which includes the models with single sources (GelSight sensor or external camera), and demonstrate that tactile and vision information are complementary to each other in the slip detection task. Especially for the objects with slippery and smooth surfaces, vision provides more cues than tactile. This work can be helpful to the area of automatic force adjustments, grasping strategy selection and fine manipulations. In addition, it also provides an example of a learning method to do multiple sensor fusion. 
\addtolength{\textheight}{-12cm}   




\section*{ACKNOWLEDGMENT}
The work is supported by Toyota Research Institute and Lincoln Laboratories. We thank Yuchen Mo and Shaoxiong Wang for helpful discussion. We thank Wen Xiong, Susan Spilecki and Pamela Siska for revising the manuscript. 


\bibliographystyle{IEEEtran}
\bibliography{Ref}

\end{document}